\documentclass{article}

     \usepackage[final,nonatbib]{neurips_2020}

\usepackage[utf8]{inputenc} 
\usepackage[T1]{fontenc}    
\usepackage{hyperref}       
\usepackage{url}            
\usepackage{booktabs}       
\usepackage{amsfonts}       
\usepackage{nicefrac}       
\usepackage{microtype}      
\usepackage{cite}
\usepackage{amsmath,amssymb,amsfonts}
\usepackage{graphicx}
\usepackage{textcomp}

\usepackage{algorithm}
\usepackage{algpseudocode}
\usepackage[misc,geometry]{ifsym}
\usepackage{mathtools}
\usepackage{graphicx}
\usepackage{color}
\usepackage{multirow}
\usepackage{array}
\usepackage{booktabs}
\usepackage{url}
\usepackage{subfig}
\usepackage[misc,geometry]{ifsym}

\usepackage[table]{xcolor}
\definecolor{rcColor}{rgb}{0.96,0.93,0.93}
\definecolor{cellColor}{rgb}{0.96,0.94,0.94}

\title{Generative Adversarial Stacked Autoencoders}

\author{
  Ariel Ruiz-Garcia\thanks{ariel.9arcia@gmail.com. Most of this work was done while at Coventry University.} \\
  LatinX in AI and SeeChange.ai\\
  Manchester, United Kingdom\\

  \And
  Ibrahim Almakky\thanks{ibrahimalmakky@gmail.com}, Vasile Palade\thanks{vasile.palade@convetry.ac.uk}, Luke Hicks\thanks{hicksl@convetry.ac.uk} \\
  Coventry University\\
  Coventry, United Kingdom\\
  
}

\def\BibTeX{{\rm B\kern-.05em{\sc i\kern-.025em b}\kern-.08em
    T\kern-.1667em\lower.7ex\hbox{E}\kern-.125emX}}
\begin{document}
 
\maketitle
  
\section{Introduction}
Generative Adversarial Networks (GANs) \cite{SCHMIDHUBER202058}, \cite{Makhzani2015} have become predominant in image generation tasks. Their success is attributed to the training regime which employs two models: a generator $G$ and discriminator $D$ that compete in a minimax zero sum game. Nonetheless, GANs are difficult to train due to their sensitivity to hyperparameter and parameter initialisation, which often leads to vanishing gradients, non-convergence, or mode collapse, where the generator is unable to create samples with different variations. 

In this work, we propose a novel Generative Adversarial Stacked Convolutional Autoencoder (GASCA) model  and a generative adversarial gradual greedy layer-wise learning algorithm designed to train Adversarial Autoencoders in an efficient and incremental manner. Our training approach produces images with significantly lower reconstruction error than {\it vanilla} joint training. 

\section{Generative Adversarial Stacked Convolutional Autoencoders}
	

	Let $x_\varphi$ be a sample from the data distribution $p_d(x_\varphi)$ and $x_\mu$ the sample from the data distribution $p_d(x_\mu)$ used as the desired target reconstruction. The autoencoder $G$ model in a GASCA learns to map $x_\varphi$ to a latent space $z$ and back to a reconstruction $y$ that resembles $x_\mu$ and lies in the distribution $q(y)$. The discriminator $D$ attempts to differentiate between $y$ and $x_\mu$. 
	
	In the conventional adversarial autoencoder framework \cite{Makhzani2015}, a distribution $p(z)$ \textemdash often a Gaussian distribution\textemdash is imposed on $q(z)$ by estimating the divergence between $q$ and $p$. This imposition can be used to produce reconstructions with specific features. However, in order to produce reconstructions that are as close as possible to the desired target image $x_\mu$, instead of imposing random noise on the hidden representation vector, the GASCA model imposes $p_d(x_\mu)$ on $q(y)$ in the following manner:    $q(y) = \mathbb{R}_{x_\mu} q(y|x_\mu)p_d(x_\varphi)dx_\mu$. With this formulation, the discriminator model $D$ is optimised to rate samples from $p_d(x_\mu)$ with a higher probability, and samples from $q(y)$ with a low probability. Formally, this is defined as: $	\nabla_{\theta_d} \frac{1}{m} \sum_{i=1}^{m}\Big[\log D(x^{(i)}_\mu) +\log \big(1 - D(G(y^{(i)}))\big)\Big] $ where $x_\mu$ is an input image and $x_\varphi$ is the target reconstruction image and $m$ is the number of samples. Note that since $(x_\varphi=x_\mu)$ is not necessarily true, the discriminator $D$ is not guaranteed to see the input to $G$. 
	
	The objective of the autoencoder model $G$ is to convince the discriminator model $D$ that a sample reconstruction $y$ was drawn from the data distribution $p_d(x_\mu)$ and not from $q(y)$. This optimisation is done according to:  
	$\nabla_{\theta_g} \frac{1}{m} \sum_{i=1}^{m} \log\Big(1 - D\big(G(y^{(i)})\big)\Big) $. Furthermore, since GANs are known to be difficult to train due to their sensitivity to hyper-parameters and parameters initialisation, which often leads to mode collapse, we propose training the GASCA model in a Greedy Layer Wise (GLW) \cite{GWL-Bengio} fashion. However, since the greedy nature of GLW leads to error accumulation as individual layers are trained and stacked \cite{Ruiz-Garcia2018d}, we build on the gradual greedy layer-wise training algorithm from \cite{Ruiz-Garcia2018d} and adapt it for adversarial autoencoders. We introduce the GAN gradual greedy layer-wise (GANGGLW) training framework and formally define it in Algorithm \ref{tab:Gradual-GLW2}.

	\begin{algorithm}
	\caption{Given a training set $X$ and validation set $\tilde{X}$ each containing input images $x_\varphi$ and target images $x_\mu$, $m$ shallow autoencoders, an unsupervised feature learning algorithm $\mathcal{L}$ which returns a trained shallow autoencoder and a discriminator model, and a fine-tuning algorithm  $\mathcal{T}$: train $D^1$ and $G^1$ jointly with raw data and add them to their corresponding stacks $G$ and $D$. For the remaining autoencoders and generator models: encode $X$ and $\tilde{X}$ using the encoder layers $\xi$ from the stack $G$. Create a new discriminator $D^k$ and train together with the new autoencoder $G^k$ and add them to their corresponding stacks. Fine-tune $G$ on raw pixel data. Forward propagate $x_\varphi \subset X$ through $G$ and use the resulting features, along with $x_\mu$, to fine-tune $D$ for binary classification.}
	\label{tab:Gradual-GLW2} 
	\begin{algorithmic}[1]
		\State $ [G^1,D^1] \leftarrow \mathcal{L} (G^1,D^1,X,\tilde{X})$
		\State $G \leftarrow G \circ G^1$
		\State $D \leftarrow D \circ D^1$
		\For{ $k=2$, \ldots, $m$ }
		\State $[\xi,\delta] \leftarrow D$
		\State $[X_g,\tilde{X}_g] \leftarrow \xi(X,\tilde{X})$
		\State $[G^k,D^k] \leftarrow \mathcal{L} (G^k,D^k,X_d,\tilde{X}_d)$
		\State $G \leftarrow G^{(k)} \circ G$
		\State $D \leftarrow D^{(k)} \circ D$
		\State $G \leftarrow \mathcal{T} (G, X,\tilde{X})$
		\State $ X_\varphi  \leftarrow G(x_\varphi)$
		\State $D \leftarrow \mathcal{T} (D, \{X_\varphi, x_\mu\subset X\} )$
		\EndFor
		\State \textbf{return}  $G,D$
	\end{algorithmic}
	\end{algorithm}

	By fine-tuning $G$ and $D$ in Algorithm \ref{tab:Gradual-GLW2} we avoid error accumulation from one layer to the next and reduce the required number of fine-tuning steps for deeper layers. 
	
	The key difference in GANGGLW as opposed to conventional GLW is that each time a new layer or shallow autoencoder is trained, the entire model is fine-tuned on raw data. This allows for stronger inter-layer connections and significantly decreased error accumulation. Moreover, both the discriminator and generator are trained in an incremental manner where layers are added one by one. 
	
    This formulation of GAN autoencoder also allows for the target reconstructions to be different than the input image as in conventional autoencoders. Figure 1 shows a sample reconstruction of facial images with different poses (up to $60$ degrees), where the same image (at $0$ degrees, top right) is used as target reconstruction. Our GANGGLW training approach has also proven to be effective when used a as a pre-training approach for supervised tasks such as classification, where it produces state-of-the-art accuracy. An extended version of this work has been published in \cite{GASCA}.

    	\begin{figure}[!htbtp]
    	\centering
    	\subfloat {{\includegraphics[width=1.55cm]{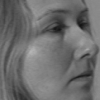} }}%
    	\subfloat{{\includegraphics[width=1.55cm]{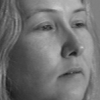} }}%
    	\subfloat{{\includegraphics[width=1.55cm]{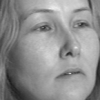} }}%
    	\subfloat{{\includegraphics[width=1.55cm]{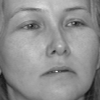} }}%
    	\subfloat{{\includegraphics[width=1.55cm]{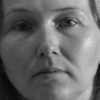} }}%
    	\\
    	\subfloat {{\includegraphics[width=1.55cm]{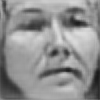} }}%
    	\subfloat{{\includegraphics[width=1.55cm]{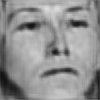} }}%
    	\subfloat{{\includegraphics[width=1.55cm]{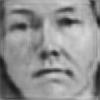} }}%
    	\subfloat{{\includegraphics[width=1.55cm]{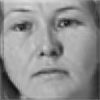} }}%
    	\subfloat{{\includegraphics[width=1.55cm]{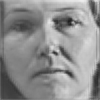} }}%
    	\\
    	\caption[GASCA$_2$ reconstructions as $0$ degrees]{Sample reconstructions produced by a GASCA model on face images from \cite{multipie}.}%
    	\label{fig:gasca_reconstructions}%
    \end{figure}

\renewcommand{\refname}{List of References}
\bibliographystyle{ieeetr} 
\bibliography{neurips_2020} 

\end{document}